\documentclass{article}

\usepackage{arxiv}

\usepackage[utf8]{inputenc} 
\usepackage[T1]{fontenc}    
\usepackage{hyperref}       
\usepackage{url}            
\usepackage{booktabs}       
\usepackage{amsfonts}       
\usepackage{nicefrac}       
\usepackage{microtype}      
\usepackage{lipsum}		
\usepackage{graphicx}
\usepackage{doi}
\usepackage{longtable}

\usepackage{pifont}

\hypersetup{
pdftitle={Architext: Language-Driven Generative Architecture Design},
pdfsubject={q-bio.NC, q-bio.QM},
pdfauthor={Theodoros Galanos, Antonios Liapis, Georgios N. Yannakakis},
pdfkeywords={First keyword, Second keyword, More},
}

\begin{document}
\title{Architext: Language-Driven Generative Architecture Design}

\author{Theodoros Galanos\thanks{Additional affiliation: Aurecon} \\
	Institute of Digital Games\\
	University of Malta\\
	Msida, Malta\\
	\texttt{theodoros.galanos.21@um.edu.mt} \\
	\And
	\href{https://orcid.org/0000-0001-5554-1961}{\includegraphics[scale=0.06]{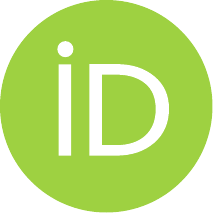}\hspace{1mm}Antonios Liapis} \\
	Institute of Digital Games\\
	University of Malta\\
	Msida, Malta\\
	\texttt{antonios.liapis@um.edu.mt} \\
	\And
	\href{https://orcid.org/0000-0003-2728-4026}{\includegraphics[scale=0.06]{orcid.pdf}\hspace{1mm}Georgios N. Yannakakis} \\
	Institute of Digital Games\\
	University of Malta\\
	Msida, Malta\\
	\texttt{georgios.yannakakis@um.edu.mt} \\
}

\maketitle

\begin{abstract}
Architectural design is a highly complex practice that involves a wide diversity of disciplines, technologies, proprietary design software, expertise, and an almost infinite number of constraints, across a vast array of design tasks. Enabling intuitive, accessible, and scalable design processes is an important step towards performance-driven and sustainable design for all. To that end, we introduce \emph{Architext}, a novel semantic generation assistive tool. Architext enables design generation with only natural language prompts, given to large-scale Language Models, as input. We conduct a thorough quantitative evaluation of Architext's downstream task performance, focusing on semantic accuracy and diversity for a number of pre-trained language models ranging from 120 million to 6 billion parameters. Architext models are able to learn the specific design task, generating valid residential layouts at a near 100\% rate. Accuracy shows great improvement when scaling the models, with the largest model (GPT-J) yielding impressive accuracy ranging between 25\% to over 80\% for different prompt categories. We open source the finetuned Architext models and our synthetic dataset, hoping to inspire experimentation in this exciting area of design research.
\end{abstract}


\section{Introduction}\label{sec:introduction}

Parametric and generative design were first introduced when numerical optimization methods and computer software were introduced to architectural design in the 1970s \cite{gero1975} and have been central to architectural practice for the past three decades \cite{CAETANO2020287}. Parametric design is typically defined as the use and study of computational representations of geometric relationships \cite{Kalay1989ModelingOA}, while generative design (GD) is defined as the process of exploring ways to generate complex forms from simple specifications \cite{Mccormack2005GenerativeDA}, typically as a means to generate solutions to a specific problem \cite{mitchell1975}. 

\begin{figure}
    \centering
    \includegraphics[width=\textwidth]{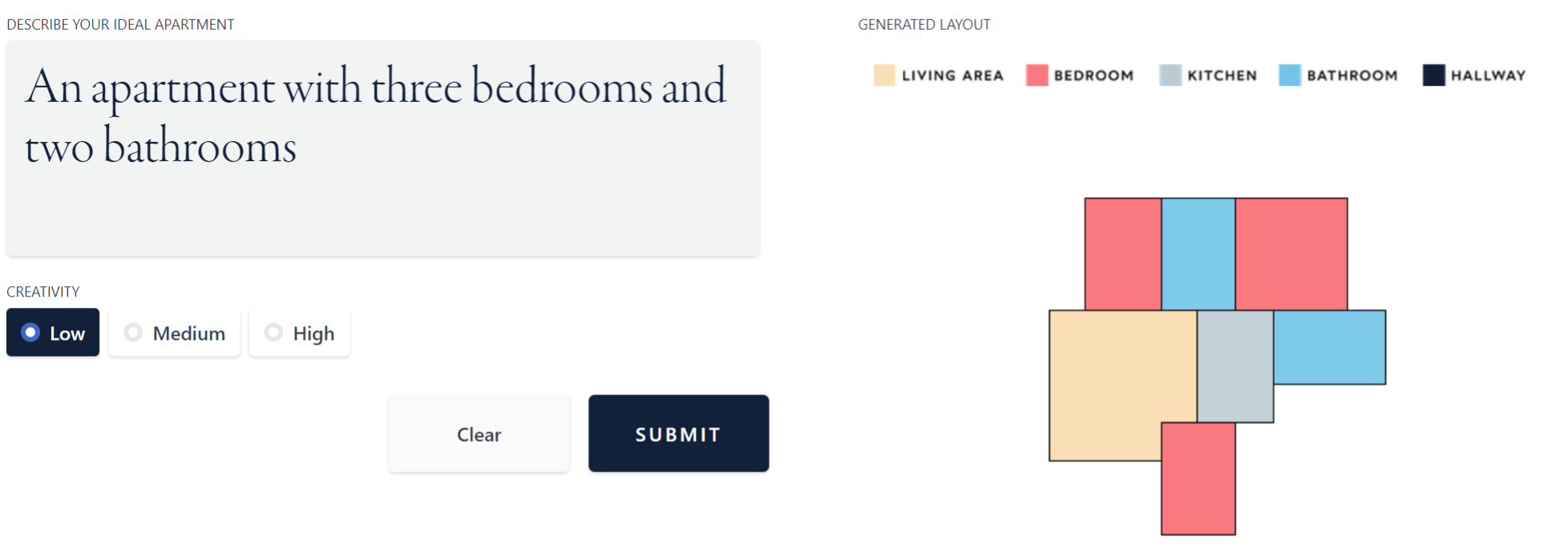}
    \caption{The user interface of Architext, as seen from the point of view of the human designer. Architext only requires a natural language prompt and outputs a layout design in different formats allowing direct visual inspection but also offering links to other design software.\label{fig:teaser}}
\end{figure}

Over the years, a number of different GD approaches have been developed such as utilizing syntactic rules \cite{stiny1985, Schirmer2011UsingSG}, expert systems \cite{flemming1988}, bubble diagrams \cite{Merrell2010}, evolutionary methods \cite{Wang2020, BuHamdan2020} and more recently, Quality Diversity evolution \cite{Galanos2021ARCHElitesQF, Gaier2018DataEfficientDE} and neural networks \cite{Azizi2020, Oh2019GenerativeDE, Nauata2020HouseGANRG, Li2019LayoutGANGG}. The majority of design exploration in practice, however, still operates within popular virtual algorithmic platforms like Grasshopper3D\footnote{Grasshopper is a visual programming language and environment that runs within the Rhinoceros 3D computer-aided design application.} and Dynamo, using specialized components and plugins developed to facilitate GD\footnote{Dynamo is a visual programming tool that gives users the ability to visually script behavior, define custom pieces of logic, and script using various textual programming languages. It is mainly used within Autodesk Revit.}. 

In recent years, large-scale Pre-trained Language Models (PLMs) have achieved great success in various natural language understanding and generation tasks \cite{Rae2021ScalingLM, Chowdhery2022PaLMSL, Thoppilan2022LaMDALM, Brown2020LanguageMA, Liu2022FewShotPF, Devlin2019BERTPO}. Their success eventually led to exploration of multi-modal settings in which language is one of the input modalities. Visual-Language Models (VLMs), for instance, have shown great potential across dissimilar applications like image generation, vision-language understanding, visual question answering, image captioning, and retrieval---see \cite{Eichenberg2021MAGMAM,Zeng2022SocraticMC} among many.

In this paper we evaluate the potential of PLMs as engines of generative architectural design. We argue that semantic generation workflows bring forth a new era of generative design that promises to alleviate the limitations of contemporary methods. For that purpose we introduce \emph{Architext}, a semantic generation system for architectural layouts that provides a flexible, open-source, and scalable method of generating a large diversity of designs (see Figure \ref{fig:teaser}). Specifically, we evaluate the capacity of a number of state-of-the-art PLMs to learn to generate semantic-based representations of architectural geometry when given only a natural language prompt describing the properties of the requested design. To accomplish this, we fine-tune these models on a synthetically generated dataset of residential layouts along with natural language descriptions of their spatial and geometric properties. We observe that when fine-tuned, PLMs are able to always generate valid geometric layouts despite using solely semantic representations of geometry without any visual input. 

In our experiments, we evaluate these models across 56 natural language prompts which express a variety of common design constraints. Results indicate that all models learn how to create valid designs at, or above, a 99\% rate. This is a clear indication of the capacity of language models to be repurposed for structured generation of geometric configurations. Meanwhile the correspondence between natural language prompts and output geometry--which we name \emph{correctness}---varies substantially across models and type of prompts. In particular, correctness seems to be heavily influenced by model size and capacity. Our highest performing model---a 6B parameter GPT-J---yields correctness scores that lie between 30\% and 83\% across the different prompt categories.

This paper is novel in a number of ways. First, we introduce Architext and through it we show that it is possible to adapt PLMs into generators of architectural designs which offer promising alternatives to current generative design workflows. Second, we develop and share publicly a synthetic dataset of architectural layouts and their semantic annotations, along with the fine-tuned models. Finally, our experiments show that the fine-tuned PLMs are able to generate valid and correct outputs at highly satisfactory levels, making them viable candidate solutions for design exploration and optimization workflows.

\section{Background}\label{sec:background}

In this section we cover the necessary background in semantic generation from the perspective of AI (see Section \ref{sec:AI_semantic}) and architecture (see Section \ref{sec:Arch_semantic}).

\subsection{AI and Semantic Generation \label{sec:AI_semantic}}

Architext builds on fine-tuned PLMs and hence its conception is tightly connected to the semantic revolution that took place within AI research community over the last year or so. We trace the beginning of this turn towards language as a fundamental modality for multi-purpose generative models in CLIP \cite{Radford2021LearningTV}. CLIP showed the potential of language-guided generation across a diverse set of tasks. Soon after its release, it inspired a range of multi-modal generative models that utilized language as the means to guide generation of different outputs. The range of applications of these models has been staggering, with ``guiding models'' used for text-to-image generation \cite{Crowson2022VQGANCLIPOD, Ding2022CogView2FA, Ramesh2022HierarchicalTI}, image editing and in-painting \cite{Gafni2022MakeASceneST, Ni2022NWALIPLG, Nichol2021GLIDETP}, image captioning \cite{Yu2022CoCaCC, Li2022BLIPBL} and segmentation \cite{Xu2022GroupViTSS, Li2022LanguagedrivenSS}, layout-based synthesis \cite{Wang2022InteractiveIS}, 3D shape generation \cite{Liu2022TowardsIT, Sanghi2021CLIPForgeTZ}, navigation \cite{Moudgil2021SOATAS, Chen2021HistoryAM}, text generation \cite{Ali2021TellingCS}, reasoning \cite{Wang2022VisuallyAugmentedLM}, visual question answering \cite{Changpinyo2022AllYM, Alayrac2022FlamingoAV}, model distillation \cite{Wang2022CLIPTDCT}, video summarization \cite{Narasimhan2021CLIPItLV}, multi-modal generation \cite{Dai2022OneMM, Su2022LanguageMC, Aghajanyan2022CM3AC}, biology \cite{Chowdhury2021, Madani2021, Madani2020ProGenLM, Rothchild2021C5T5CG} and reinforcement learning and planning \cite{Yang2021SafeRL,Mirchandani2021ELLAET,Ahn2022DoAI,Rahtz2022SafeDR,Sharma2022CorrectingRP,Tam2022SemanticEF,Liu2022AskingFK}.

Research on code generation is perhaps the closest to the task of generative architectural design by natural language prompts. Layouts represented semantically as geometric entities present some similarities to code, especially with respect to how sequences are structured along with the logic and constraints that guide that structure. Xu et al. \cite{xu2022ide} consider the impact of semantic generation models on the actual professional workflows. While Architext was not designed in this way, the implications of having a neural network model that turns a (natural language) concept into a layout---and the difficulties of doing this in traditional workflows---were the key motivations behind its design. Guo et al. \cite{Guo2021LearningTC} generated code completions where the model learns to insert special tokens in places of high uncertainty, enabling a more involved HCI between the model and the developer. Nye et al. \cite{Nye2021ShowYW} attempted to improve language model performance in scenarios where multi-step computations are required to generate a solution. They find that by representing the problem as an algorithmic task, and encoding the intermediate steps as text, improves the models' performance substantially. Architext instead learns to generate layouts as one coherent entity composed of different rooms. Finally, Poesia et al. \cite{Poesia2022SynchromeshRC} combine retrieval and sampling to improve code generation from PLMs. Using a specialised module, they enforce a number of different design constraints in the model generations. Architext on the other hand learns different design constraints simply from the data, using text as the means to communicate these constraints to the model. This simplifies the process and allows for an easier way of integrating new constraints and better interpretability of the model's performance, from the point of view of the human designer. 

\subsection{Semantic Generation in Architecture\label{sec:Arch_semantic}}

Following the explosion of semantic generation and multi-modal models, there has been substantial research on different methods for generating a variety of architectural outputs. Kuo et al. \cite{Kuo2021Patch2CADPE} tackle the problem of 2D image to 3D shape translation, leveraging existing CAD databases to train a model that constructs a joint embedding space between the 2D and 3D representations of designs. The semantics utilized in that study are visual semantics---at the level of patches within the images---enabling part similarity reasoning as well as retrieval and understanding of objects' shapes within an image. Architext, on the other hand, does not operate on the visual domain at all, abstracting the complexity of the process through codified shape descriptions. 

Ganin et al. \cite{Ganin2021ComputerAidedDA} propose a method of automatically generating CAD sketches by combining a language model and an off-the-shelf serialization protocol used to describe structured objects. They collect a substantial training dataset of 4.7M parametric CAD sketches. Their model has the ability for both unconditional and conditional sketch generation, the latter made possible by feeding an embedding of a visual render of the type of sketch required as an input to the model. Similarly, Para et al. \cite{Para2021SketchGenGC} propose a generative model that is trained on a carefully designed sequential language for primitives and constraints involved in typical CAD sketches. They define a formal language for CAD sketches, with its own grammar and production rules. In contrast to these studies, Architext allows the generation of a rich and diverse set of outputs which rely only on natural language prompts provided by the human designer as the conditioning signal. While layouts are represented as simple geometric primitives (series of coordinates and space labels), the constraints involved in the generation process are also embedded using natural language. This approach simplifies the process of data generation, allows for much more flexible training, makes adding new constraints to models straightforward, and provides a clear `interface' for model interpretability.  

Similarly to the aforementioned studies, Paschalidou et al. \cite{Paschalidou2021ATISSAT} design an auto-regressive architecture with the goal of creating realistic and diverse indoor design environments. Just like Architext represents architectural layout design as a sequence generation, they represent scene synthesis as sequence generation and train their model using only labeled 3D bounding boxes. Architext focuses on the generation of layouts rather than the locations of indoor objects---which present their own set of constraints and challanges---while also taking advantage of PLMs to achieve a high diversity of outputs. 

\section{Methodology}\label{sec:methodology}

\begin{figure}
    \centering
    \includegraphics[width=15cm]{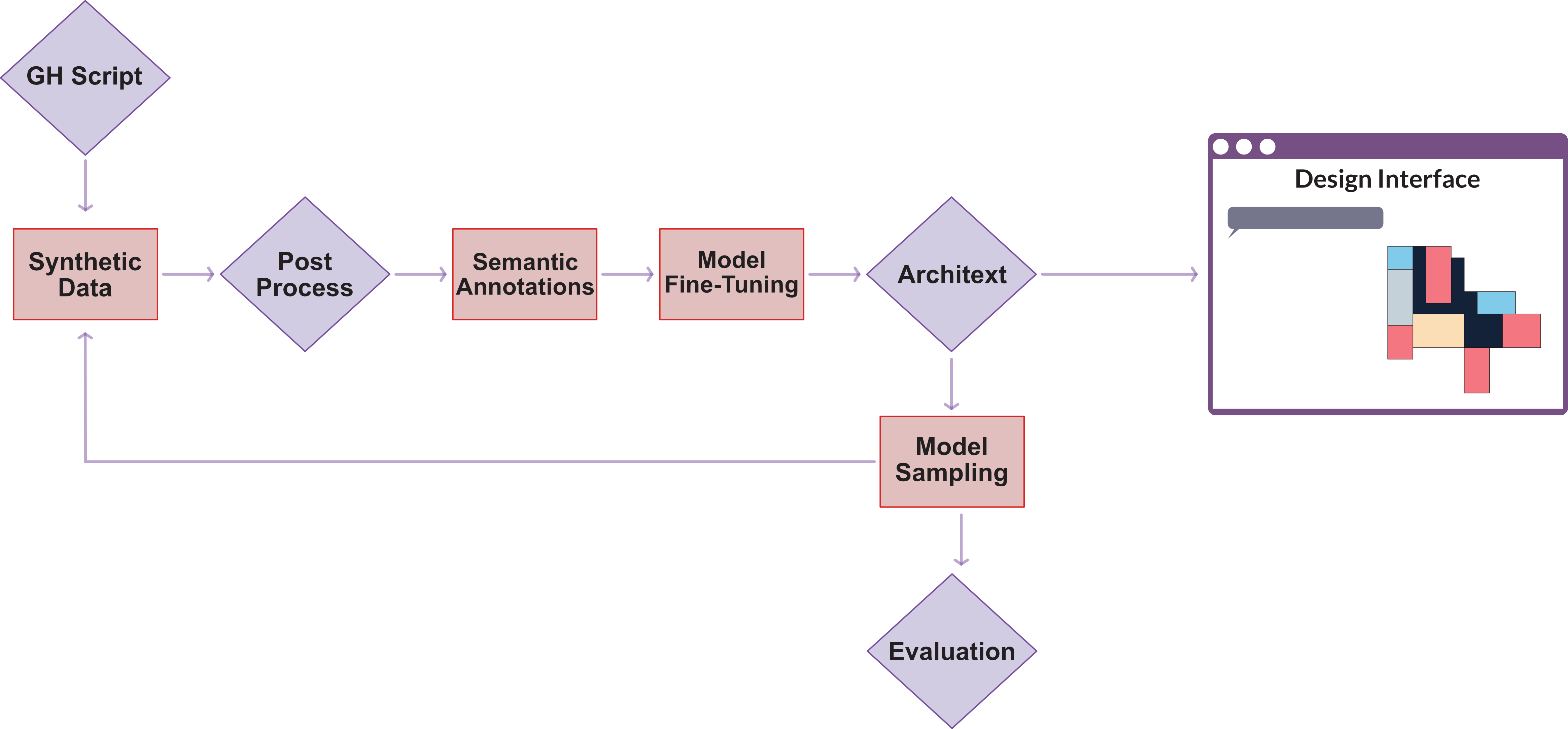}
    \caption{Summary of our data generation and model training workflow.} \label{fig:method}
\end{figure}

The goal of this study is three-fold: (1) to enable the generation of a large variety of residential floor plans with the use of fine-tuned PLMs; (2) to evaluate the performance of these models across different metrics; and (3) to assess the impact of model properties, such as scale and pre-training, on model performance. For this purpose, each floor plan is represented semantically by an $\{annotation, geometry\}$ tuple\footnote{This representation is not specific to residential floor plans; it can be used for almost any design task.} and is fed to a PLM during finetuning (see Figure \ref{fig:method}). 

By translating floor plans to language, the task of design generation becomes that of language generation, an area where a number of powerful, open-source PLMs have been used successfully in recent years. Fine-tuning these PLMs on semantic representations of floor plans is the underlying method featured in {Architext}. The PLMs we rely upon in this work are GPT-2 \cite{radford2019rewon}, GPT-Neo \cite{gptneo, gao2020pile}, and GPT-J \cite{gpt-j}.

\subsection{Synthetic Dataset Generation and Pre-processing}\label{sec:methodology_synthetic}

Architectural floor plan datasets are not generally available, certainly not at the scale required for language model training. Additionally, the datasets that are available are typically in formats that are difficult to process (e.g. PDF drawings) and need to be vectorized to be useful. This makes creating a consistent, and coherent, large-scale dataset a difficult endeavor. Even if we somehow managed to create such a dataset from the scarce sources available, the low quality and noisy data would have a detrimental impact on model performance.

To resolve this issue, we instead created a synthetic dataset using the virtual programming interface of Grasshopper 3D. Grasshopper is a plugin of the 3D modelling software Rhinoceros 3D \cite{mcneel-rhinoceros}, which is a popular software within the architectural design community. A parametric design script was developed to generate a large number of residential floor plans, across different categories. We used the Magnetizing Floor Plan Generator plugin \cite{magnetizing}, a tool that enables the generation of such designs with simple and straightforward input parameters like room types, floor area, connectivity graph, and an entrance location. During generation, we permuted the area and connectivity inputs in order to increase the diversity of the synthetic dataset. {Five different types of rooms could be placed in the layouts: bathroom, bedroom, kitchen, living room, corridor}. In total, $250,000$ unique residential floor plans were created across 6 categories. These categories were classified according to the available number of bedrooms and bathrooms, respectively: 1/1, 2/1, 2/2, 3/2, 4/2, and 4/3.

\begin{figure}
    \centering
    \includegraphics[width=\textwidth, height=20cm, keepaspectratio]{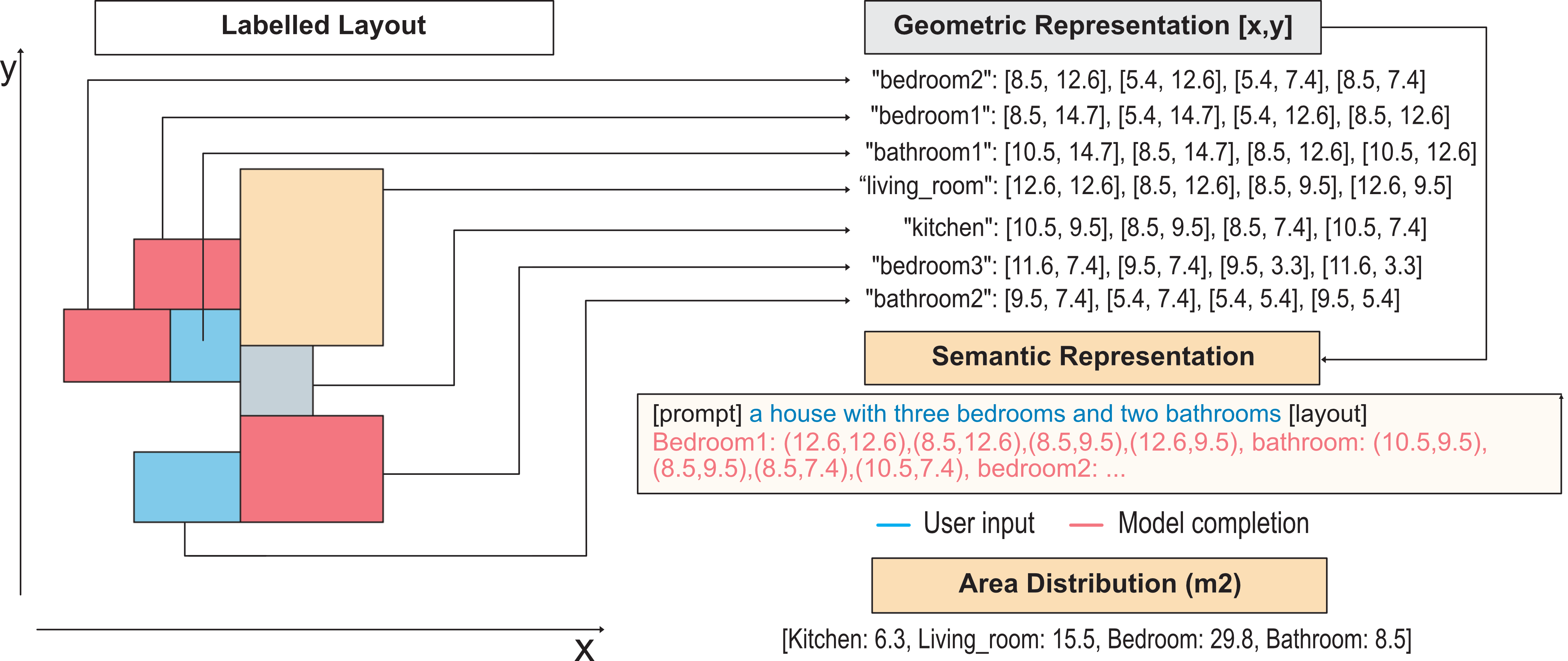}
    \caption{Semantic representations of a generated layout, extracted using its geometric properties (i.e. room types and coordinates).\label{fig:representation}}
\end{figure}

Specific care was taken to export the floor plans into an appropriate format for training. Aside from a visual representation of the layout, the spatial and geometric properties of each design were also exported as arrays. This language-based geometric representation included the labels for each room along with its coordinates, in the form of ($x, y$) coordinate tuples (see Figure \ref{fig:representation}). Given that all rooms, with the exception of corridors, were constrained to rectangular shapes, we could describe each room with four such tuples representing its corners, e.g. $bedroom: (13,12),(8,12),(8,9),(13,9)$. Each floor plan was scaled to fit within the $(0,0), (256,256)$ bounding box in order to easily couple semantic representations with floor plan images. This representation has certain advantages: (1) it offers a good compromise between abstraction and specificity, (2) it allows us to synthetically generate more designs by intervening on the geometry itself, and (3) it expresses designs using language which allows us to take advantage of pre-trained language models.

\subsection{Extracting Semantic Annotations}
As discussed, we chose to represent each floor plan with an $\{annotation, geometry\}$ tuple. \emph{Geometry} is offered by the language-based geometric representation we detailed earlier. \emph{Annotation} offers the natural language description of each geometry. To generate such a description we used the topological properties and category attributes of each floor plan as follows:

\begin{itemize}
    \item \textbf{Room number (RG)}: annotations that include information concerning the total number of rooms in each floor plan. \\
    Example: \textit{``a house with five rooms''}.
    \item \textbf{Room labels (RS)}: annotations that include information concerning the number of rooms, specifically those of the bedrooms and bathrooms, in each floor plan. \\
    Example: \textit{``a house with two bedrooms and three bathrooms''}.
    \item \textbf{Room adjacency (AP, AN)}: annotations that include information concerning adjacency between different rooms in each floor plan. Annotations included both positive adjacency (AP) and negative adjacency (AN) between different types of spaces. \\
    Examples: \textit{``the living room \textbf{is} adjacent to the bedroom''} (AP), \textit{``the kitchen \textbf{is not} adjacent to the bathroom''} (AN).
    \item \textbf{Room location (LU, LNU)}: annotations that include information concerning location of specific rooms within each layout. Eight different locations were used, following the compass: North, North East, East, South East, South, South West, West, and North West. Each floor plan was assumed to be in the same orientation, with North being `up'. Finally, a semantic distinction was made for annotations with a unique room in a specific location of the layout (LU) versus one of many rooms (LNU). \\
    Examples: \textit{``\textbf{a} bedroom is in the north side of the house''} (LNU), \textit{``\textbf{the} bathroom is in the south east side of the house''} (LU)
\end{itemize}

A custom made script was used to extract these annotations. Care was taken to shuffle and remove duplicate annotations at the floor plan level. The exact number of annotations for each floor plan varied according to the number of rooms and spatial characteristics. In total, $5,145,750$ annotations were generated for the $250,000$ synthetic layouts. These provided the required semantic information to create a complete language-based description of each floor plan, one that could be now used to fine-tune PLMs. Each entry of the final dataset is a tuple with a prompt and layout (see Fig.~\ref{fig:method} for an example). 

\subsection{Model Fine-tuning}

We fine-tune a range of models across 3 different architectures (GPT-2, GPT-J and GPT-Neo) and across different context length sizes. For GPT-2 models, we used 512 and 1024 tokens, while for GPT-Neo we used 512, 1024, and 2048 tokens. Each model is fine-tuned for 3 epochs with a batch size 32 and 4 steps of gradient accumulation. We use the adam optimizer, with an initial learning rate of $5\cdot10^{-5}$, annealed to a final learning rate of $5\cdot10^{-6}$, a weight decay of 0.1, and a warmup period of 10\% of the total epoch steps. For GPT-J, all fine-tuning was done in the original 2048 context length. We observed performance deterioration when fine-tuning for more than 1 epoch, hence we limit the process to slightly less than one epoch. The batch size and the steps of gradient accumulation used for GPT-J training were 8 and 32, respectively.

\subsection{Floor Plan Generation}

In order to evaluate model performance on the downstream task of generating floor plans, we constructed a series of language prompts that were fed to each fine-tuned model. We used 58 prompts in total and generated layouts for each prompt across different sampling parameters. This process resulted in approximately $235,000$ generated floor plans for each fine-tuned model. The evaluation process for each model output is described in the next section.

\section{Evaluation Framework}

Simulating and evaluating open-ended tasks that resemble human activities---such as architectural design---is not trivial and generally requires human evaluation of generated outputs. To bypass this issue, we focus on evaluating the generation capacity of the fine-tuned models by providing them with a set of specific, natural language prompts and conduct an analytical evaluation of their continuations according to different metrics.

\subsection{Metrics}\label{sec:metrics}

We investigate three aspects of generation performance: \emph{validity}, \emph{correctness} and \emph{diversity}. A layout that is composed of valid geometric elements, and relationships between them, is considered valid. A valid layout, whose structure corresponds to the semantic content of the prompt, is considered correct. Finally, we quantify the distribution of generated layouts, along with out-of-distribution generation, for each prompt to assess the model's generative diversity. In the remainder of this section we detail the calculation of each one of the four metrics considered.

\textbf{Validity} relates to the geometric properties of each layout. A valid layout should consist of a number of closed polygons, and polygons can have adjacent edges but must be non-intersecting. Finally, no `orphan' polygons are allowed, i.e. polygons that have no adjacent geometry. To perform a consistent evaluation of validity, we develop a post-processing script that takes a model output (natural language description) and transforms it into a geometric layout, evaluating it against the above requirements. We apply the same script to all models and consider invalid any generation that fails to be parsed by the script. 

\textbf{Correctness} evaluates whether the generated layout captures the semantic content of the user's prompt. We evaluate models across all properties provided during training: number and type of each space, adjacency, and location. {This is achieved by running the same process that generated the semantic annotations for our synthetic data. For each generated layout, the user's prompt existence is checked against all the possible annotations for that layout. If the user's prompt exists in the possible annotations, the layout is considered semantically correct.}

\textbf{Diversity} of generated layouts is important, as the model should be able to generate a variety of different designs. Diversity in generated output is not only an evidence that the model does not overfit to training data, but is also an important property in many potential downstream workflows and tasks (e.g. using the model as a generator for population-based search). We use two dimensions of diversity, both of which assess the difference of generated output from the training data (rather than from other generated content). We evaluate two different types (or dimensions) of diversity: 
\begin{itemize}
    \item \textit{Out-of-distribution (OOD) ratio} measures the percentage of generated designs that do not belong to a class of residential floor plans in the synthetic training data. As noted in Section \ref{sec:methodology_synthetic}, {the training data includes 6 categories based on number of bathrooms and bedrooms, and thus layouts with a different combination of number of bedrooms and bathrooms than those six is considered OOD}. Diversity in this dimension is measured as the ratio of OOD layouts to the total number of generated layouts per prompt.
    \item \textit{Spatial diversity} compares the total floor area of different types of spaces within a generated floor plan from the average distribution of respective types' total floor areas within the training data. We define spatial diversity as the earth mover’s distance between the histograms of these two distributions. {Spatial diversity indicates the capacity of the models for diverse design generation, an important characteristic if we want these models to be used for generative design in practice.}
    
\end{itemize}

\subsection{Sampling from fine-tuned models}

We used the nucleus sampling implementation \cite{holtzman2020curious} as provided in the HuggingFace transformer library \cite{wolf-etal-2020-transformers} to sample from all fine-tuned models. For the GPT-J models, we used the provided scripts in the mesh-transformer-jax library \cite{mesh-transformer-jax} to convert them to a PyTorch model before sampling with the same process. Each model was sampled across the 58 natural language prompts found in \ref{sec:appendix}.

\section{Results}

In this section we go through three key research questions set in this paper and address them empirically through the corresponding three metrics (validity, correctness and diversity) introduced earlier. In particular, we first examine whether language models can represent and generate valid residential floor plans (Section \ref{sec:results_LM}), we then evaluate the accuracy of PLM-based generation (Section \ref{sec:results_accuracy}) and finally we investigate how diversity alters across layout types and scales (Section \ref{sec:results_diversity}).

\begin{table}
\centering
\caption{Average validity across different prompt categories. Highest values per prompt category are shown in bold.}\label{tab:validity}
\begin{tabular}{r*7c}
\toprule
Model &  \multicolumn{6}{c}{Average Validity (\%)} & Parameters (millions)\\
\midrule
{} & RG & RS & AP & AN & LU & LNU & {}\\
\midrule
GPT-Neo-Small &  99.59 & 99.56 & 99.58 & 99.61 & 99.62 & 99.67 & 125\\
GPT2-Small   &  99.87 & 99.93 & 99.90 & 99.93 & 99.92 & 99.85 & 125\\
GPT2-Medium  &  99.91 & 99.96 & 99.96 & 99.97 & 99.94 & 99.94 & 355\\
GPTJ-6B  &  \textbf{100} & \textbf{100} & \textbf{100} & \textbf{100} & \textbf{100} & \textbf{100} & 6000\\
\midrule
Samples &  600 & 1200 & 800 & 800 & 1600 & 800 & \\
\bottomrule
\multicolumn{8}{p{14.5cm}}{RG: Rooms, General; RS: Rooms, Specific; AP: Adjacency, Positive; AN: Adjacency, Negative; LU: Location, Unique; LNU: Location, Non-unique.}\\
\multicolumn{8}{p{12.5cm}}{}
\end{tabular}
\end{table}

\subsection{Validity: Can Language Models Learn to Represent and Generate Valid Layouts? \label{sec:results_LM}}

The first step towards semantic generation of structured outputs using PLMs is the ability to fine-tune such models to consistently generate such outputs, with the required attributes. As detailed earlier a valid layout consists of a collection of adjacent, closed polygons. The models have to show the capacity of consistently generating outputs of this kind, across all the provided prompts. 

The results depicted in Table \ref{tab:validity} indicate that all models are robust and are able to consistently learn how to generate valid outputs, across all the prompts they were given. In particular, validity reaches at least 99.5\%, with performance increasing slightly across model size. Being able to fine-tune large-scale PLMs that are able to generate valid layout generators is a key outcome of this research, given the complexity of developing generative design workflows which typically require domain expertise and the use of specialized design software. These models represent an easy to use and deploy generative design program which can be utilized in a variety of design workflows. We would note here that valid layouts are provided via the post-processing step which

\subsection{Correctness: How Accurate are the Language Model Generations?}
\label{sec:results_accuracy}

\begin{table}[!tb]
\centering
\caption{Average correctness across different prompt categories. Highest values per prompt category are shown in bold.}\label{tab:accuracy}
\begin{tabular}{r*7c}
\toprule
Model &  \multicolumn{6}{c}{Average Correctness (\%)} & Parameters (millions)\\
\midrule
{} & RG & RS & AP & AN & LU & LNU & {}\\
\midrule
GPT-Neo-Small &  15.26 & 7.75 & 61.53 & 60.58 & 0.94 & 16.97 & 125\\
GPT2-Small   &  15.13 & 7.75 & 63.06 & \textbf{60.60} & 0.98 & 16.99 & 125\\
GPT2-Medium  &  14.92 & 7.71 & 63.02 & 59.89 & 1.01 & 17.21 & 355\\
GPTJ-6B  &  \textbf{83.73} & \textbf{51.81} & \textbf{71.58} & {56.61} & \textbf{25.00} & \textbf{42.06} & 6000\\
\midrule
Samples &  600 & 1200 & 800 & 800 & 1600 & 800 & \\
\bottomrule
\multicolumn{8}{p{14.5cm}}{RG: Rooms, General; RS: Rooms, Specific; AP: Adjacency, Positive; AN: Adjacency, Negative; LU: Location, Unique; LNU: Location, Non-unique.}\\
\end{tabular}
\end{table}

While being able to consistently generate valid outputs is important, it is also crucial to be able to create outputs that accurately reflect the properties requested by the user, as those are captured in the prompts given to the LM. Table \ref{tab:accuracy} shows the ratio of correct layouts (matching the prompt) over all generated layouts, averaged per prompt category. We can observe that certain types of constraints are difficult for the smaller models to satisfy, specifically those that have to do with unique locations of space types (LU) and specific number of rooms requested for different space types (RS). The same models preform slightly better on constraints that can be satisfied by a larger variety of generated designs (RG and LNU). We also note that models can achieve high correctness for adjacency-related prompts, potentially because space adjacency can be satisfied by a broader variety of spatial configurations.

When evaluating the 6 billion parameter GPT-J model, however, we observe that it outperforms the smallest models in the categories where satisfying the constraint imposes stricter requirements to generated designs. While it performs $5\%$ less in satisfying negative adjacency constraints (AN), its correctness performance is much higher when satisfying prompts requesting for number of rooms and location of specific spaces (RG, RS). These results indicate that increasing the number of model parameters can have a significant impact on model correctness for the specific prompts and design task. This finding may be an indication of the impact of scaling in generating structured outputs with language models, reflecting other similar empirical results in the literature \cite{kaplan2020scaling, henighan2020scaling, rae2021scaling, prato2021scaling, hoffmann2022training}.

\begin{figure}
    \centering
    \includegraphics[width=\textwidth, height=20cm, keepaspectratio]{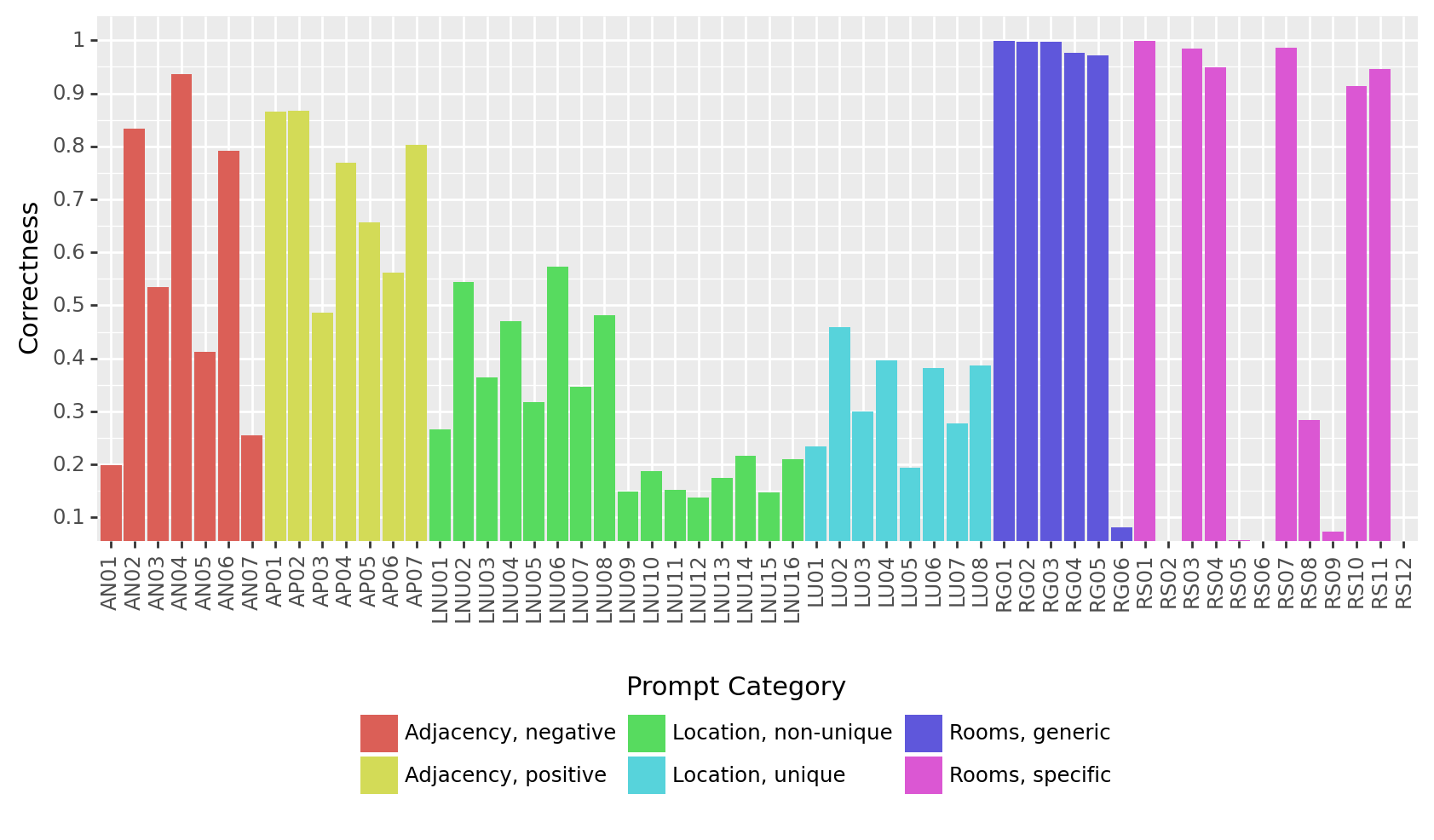}
    \caption{Correctness of generated residential floor layouts for each prompt for GPT-J. \label{fig:correctness}}
\end{figure}

Performance clearly varies across prompts of each category presented in Table \ref{tab:accuracy}. We focus on the best performing GPT-J model, and look at correctness per prompt individually in Figure \ref{fig:correctness}. For RS prompts, we observe that every prompt that describes an out-of-distribution floor layout (e.g. a house with three bedrooms and three bathrooms for RS08) has a much lower correctness value compared to in-distribution layouts. For RG prompts, this is implicitly occurring e.g. for RG06 which requests a house with ten rooms: the generated dataset could include up to 4 bedrooms and 3 bathrooms, so even with the other rooms the training data rarely reaches ten rooms. The performance of the model on the remaining categories varies less, given that a larger variety of designs can generally satisfy those adjacency and location constraints. The results illustrated in Figure \ref{fig:correctness} indicate that---at least with the current available data---extending a model's capacity to generate \textit{accurately} new, unseen designs during inference might be hard for certain layout types and that additional training data might be required for such layouts.

\subsection{Diversity: How does Diversity Change across Plans and Scales?} \label{sec:results_diversity}

Given that architectural design is fundamentally a creative process, the models' ability to generate not only accurate but also novel architectural outputs is crucial. As detailed in Section \ref{sec:metrics}, we use two different indicators to quantify diversity within each prompt category and across all models. 

\begin{table}[!tb]
\centering
\caption{Out-Of-Distribution (OOD) Ratio of Generated Layouts. Highest values per prompt category are shown in bold.} \label{tab:OOD}
\begin{tabular}{l*7c}
\toprule
Model &  \multicolumn{6}{c}{Average OOD Ratio (\%)} \\
\midrule
{} & RG & RS & AP & AN  & LU & LNU & {}\\
\midrule
GPT-Neo Small &  7.81 & 8.42 & 9.51 & 8.98 & 8.80 & 9.24 5\\
GPT2-Small    &  9.28 & 9.16 & 9.13 & 8.92 & 9.00 & 8.96 \\
GPT2-Medium   &  \textbf{9.70} & \textbf{9.48} & 9.31 & 9.27 & 9.99 & \textbf{9.42} \\
GPTJ &  0.68 & 8.85 & \textbf{21.55} & \textbf{16.36} & \textbf{24.85} & 0.68 \\
\midrule
Samples &  600 & 1200 & 800 & 800 & 1600 & 800 & \\
\bottomrule
\multicolumn{8}{p{14.5cm}}{RG: Rooms, General; RS: Rooms, Specific; AP: Adjacency, Positive; AN: Adjacency, Negative; LU: Location, Unique; LNU: Location, Non-unique.}\\

\end{tabular}
\end{table}
\begin{figure}
    \centering
    \includegraphics[width=\textwidth, height=20cm, keepaspectratio]{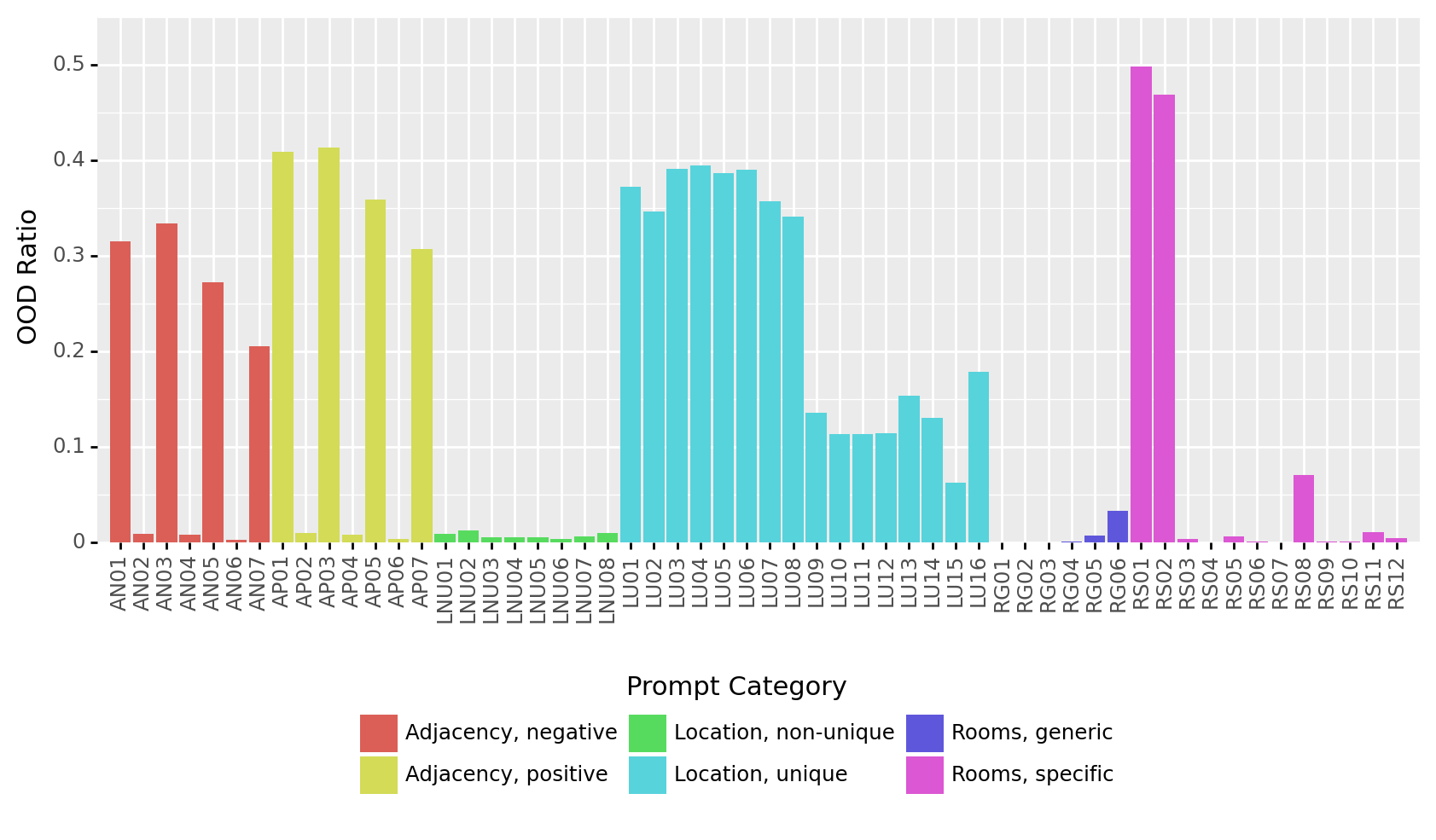}
    \caption{The OOD ratio of generated floor layouts across each prompt for GPT-J.}\label{fig:ood}
\end{figure}
Table \ref{tab:OOD} shows the out-of-distribution ratio of layouts compared to the training data, averaged per prompt category. It is evident that the smaller pre-trained models have OOD ratio values that range between $7\%$ and $9\%$ across all prompt categories. The ability to generate more diverse outputs seems to correlate with model size. The OOD ratio appears to increase in 5 out of 6 categories as model size increases. This is especially evident with the 6 billion parameter GPT-J model which has substantially higher OOD ratio values in three (AP, AN, and LU) of the six categories. Interestingly, while the model remains competitive in the 2 out of the remaining 3 categories (i.e. RS and LNU), we observe that it almost completely fails to generate out of distribution layouts for the generic room prompts (RG). This finding is perhaps an indication of model overfitting to the training data. 

The ability of GPT-J to provide such diverse outputs while achieving much higher correctness scores, as shown earlier in Table \ref{tab:accuracy}, is indicative of the effect that network size has to model performance; GPT-J has 20 times the parameters of other GPT variants. Additionally, GPT-J was trained on the Pile---an 825 GiB English text corpus targeted at training large-scale language models---constructed from 22 diverse high-quality subsets \cite{Gao2021ThePA}. The higher quality and more diverse pre-training, along with the superior architecture of the model, are also important factors in model performance and diversity.

Figure \ref{fig:ood} shows the OOD ratio of content generated by GPT-J per prompt. We note that the ability to generate OOD floor layouts varies greatly not only across different categories of prompts but also across different prompts within the same category. The only category where the model is consistently able to generate a large ratio of OOD designs is LU prompts, with half of the prompts producing OOD layouts more than 30\% of the time. For the LNU category, the model is capable of generating only a few OOD layouts across the provided prompts. We can only assume that the uniqueness constraint restricts the model's design space into floor layouts that are at the margins, or outside, the initial training data distribution (e.g. layouts with a single bedroom). 
For both adjacency constraints (AN and AP) we observe that the model can generate a high percentage of OOD floor layouts for the prompts that request an (non-)adjacency that can be satisfied with designs that contain only one space of that type (i.e. prompts of the type ``the bedroom is adjacent to\ldots''). On the other hand, the model almost completely fails to produce OOD outcomes on the prompts that allow for multiple such spaces to exist, as long as one of them is adjacent to the requested space (i.e. prompts of the type ``a bedroom is adjacent to\ldots''). Similarly to the location categories, the reason appears to be the restriction imposed by the uniqueness constraint to the model's latent space, with most typologies that satisfy it found typically at the margins, or outside, the typologies provided in the initial training data (e.g. layouts with one bedroom and multiple bathrooms).
As a general finding we believe that a strategy that could assist OOD layout generation is to prompt the model near the `boundary' of its training distribution, for example with prompts targeting designs with the least or most number of rooms in the training dataset. This helps the model access an area of the latent space that contains more OOD designs. Such an approach also comes with several challenges. As seen in Figure \ref{fig:ood}, the remaining OOD categories in the RS set fail to produce a high percentage of OOD designs. This seems to be because the model is unable to generate valid designs at that region of the design space, and instead `falling back' to the closest typologies in the training distribution to those prompts. This assumption of ours is open for further examination.

\begin{table}
\centering
\caption{Spatial diversity in generated layouts, split between in-distribution and  OOD generated content. Highest values per prompt category are shown in bold.} \label{tab:wd}
\begin{tabular}{l*7c}
\toprule
Model & \multicolumn{6}{c}{Wasserstein Distance (In$|$Out)} \\
\midrule
{} & RG & RS & AP & AN & LU & LNU & {}\\
GPT-Neo-S &  4.26$|$4.36 & \textbf{6.03}$|$5.32 & 4.28$|$4.38 & \textbf{4.52}$|$4.52 & \textbf{4.27}$|$4.39 & 4.97$|$4.76\\
GPT2-S    &  4.26$|$4.31 & 4.20$|$4.26 & 4.23$|$4.28 & 4.20$|$4.27 & 4.19$|$4.24 & 4.22$|$4.29\\
GPT2-M    &  \textbf{4.35}$|$4.38 & 4.31$|$4.34 & \textbf{4.31}$|$4.35 & 4.33$|$4.37 & 4.38$|$4.42 & 4.31$|$4.34\\
GPTJ-6    &  2.80$|$\textbf{6.03} & 2.90$|$\textbf{5.98} & 3.50$|$\textbf{6.16} & 3.85$|$\textbf{6.16} & 2.32$|$\textbf{5.46} & \textbf{5.57}$|$\textbf{6.57}\\

\bottomrule
\multicolumn{8}{p{14.5cm}}{RG: Rooms, General; RS: Rooms, Specific; AP: Adjacency, Positive; AN: Adjacency, Negative; LU: Location, Unique; LNU: Location, Non-unique.}\\
\end{tabular}
\end{table}

\begin{figure}
    \centering
    \includegraphics[width=\textwidth, width=.95\textwidth, keepaspectratio]{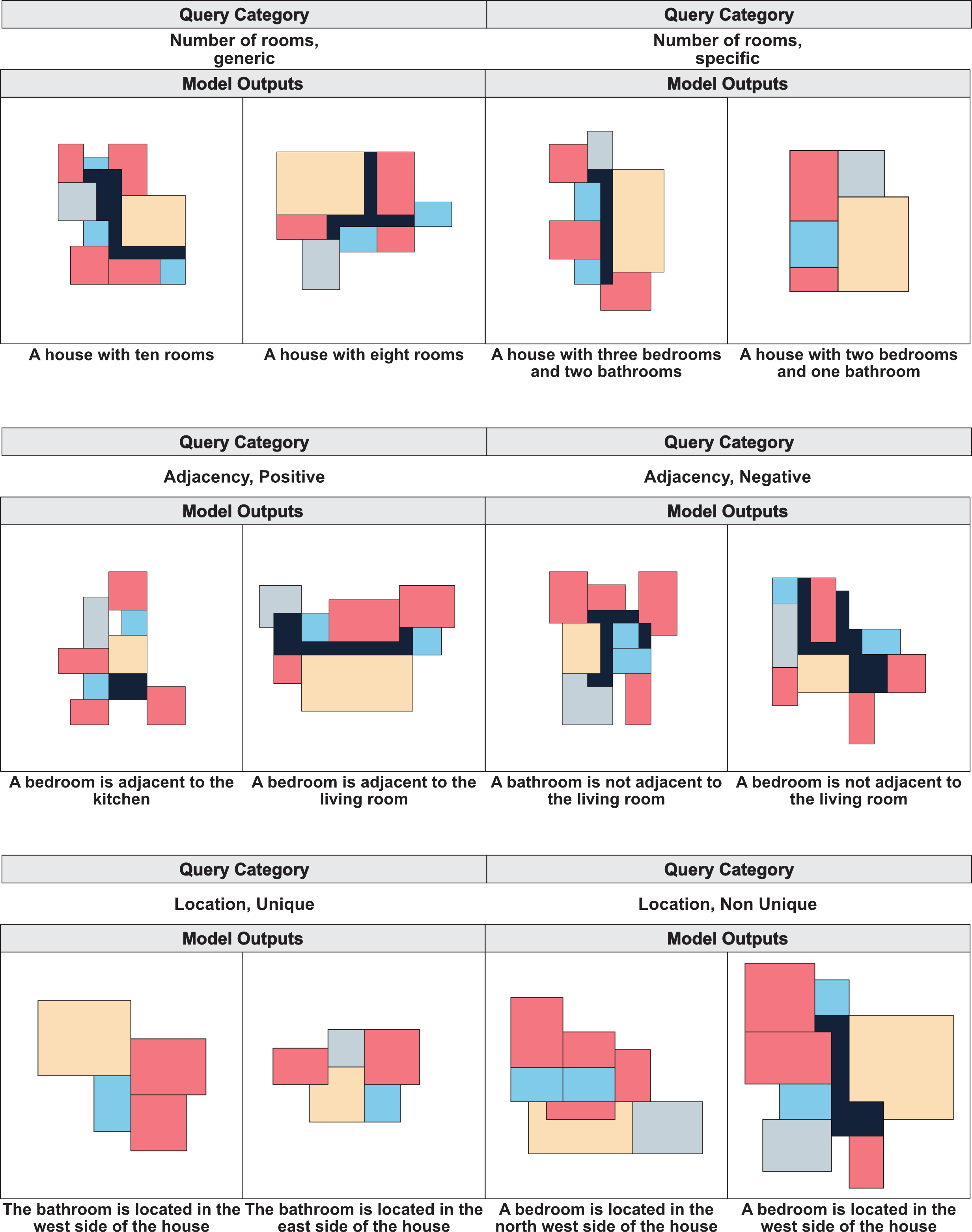}
    \caption{Valid and correct generated layouts with the highest spatial distance, per prompt category.} \label{fig:wd}
\end{figure}

Assessing diversity through the second metric detailed in Section \ref{sec:metrics}, Table \ref{tab:wd} shows the spatial diversity of each model per prompt category. We assess spatial diversity as the Wasserstein distance of the total floor area per room type of a generated layout and the average respective floor area in the training data. Since some layouts are in-distribution and thus match the training data more than OOD generations, we split spatial diversity between those layouts that match layouts in the training data and OOD generated layouts. For in-distribution layouts, GPT-Neo-Small, GPT2-Medium and GPT-J score  higher in, respectively, three, two, and one prompt categories compared to GPT-J. The three smaller models are generally very close in terms of Wasserstein distance for both in- and out-of-distribution generations and show little deviation across those distributions. On the other hand, GPT-J yields higher Wasserstein distance for out-of-distribution generations across all prompt categories, indicating that it creates more diverse designs when it is sampled in a way that promotes OOD generation (e.g. using open ended prompts). GPT-J's superior performance in terms of correctness of generated designs, along with its capacity to generate diverse outputs in OOD regions, make it a very powerful generator for downstream design exploration tasks.
A glimpse of the model's potential for generating high design diversity can be seen in Figure \ref{fig:wd}, where we visualise the two layouts with the highest Wasserstein distance for each prompt category. A number of unique and dissimilar designs emerge from Architext for each particular prompt presented.

\section{Discussion}\label{sec:discussion}

Below we outline our key findings, discuss the challenges and limitations of this work and go through a number of important future research directions for Architext.

\subsection{Summary of Results}\label{sec:results_summary}

Through experiments reported in this paper, we were able to finetune a range of pre-trained models that can produce residential layouts at a reasonable accuracy across 6 different categories of language prompts, which cover dissimilar characteristics of the design (i.e. typology, location, adjacency). All trained models---irrespective of size and model architecture---were able to create valid architectural layouts at a rate higher than 99.5\%, with the largest model (GPT-J) reaching 100\% validity across all prompt categories. 

When considering how well the generated layouts satisfy the constraints communicated in the natural language prompt, our key finding is that smaller models (GPT-Neo, GPT2) struggle with specific categories of prompts, especially prompts requesting for layouts with a specific number of rooms and prompts requesting specific locations of unique spaces within a layout. Such models however show good performance (over 60\% correctness) when prompted with adjacency constraints between spaces. We argue that better performance does not necessarily reflect model capacity but rather that particular constraints can be satisfied in more ways and layouts than others. 

While the accuracy values obtained were not as satisfactory for the smaller models, GPT-J---a much larger model with 6 billion parameters---showed much better performance across all categories. Correctness ranged from 25\% for the LU prompt category, a 25x improvement, to 83\% for the RS category, a 7x improvement, over smaller models. These results show that model scaling greatly improves model performance and, in turn, such models could assist with much more complex generative applications. Despite the increase in model size and model performance, as well as the relatively narrow task specification, GPT-J does not appear to completely overfit. Results show that GPT-J manages to generate a much larger number of out-of-distribution designs for 3 out of the 6 categories, while performing worse in 2 out of 3 categories. The combined abilities of creating valid and diverse designs highlights the potential for Architext models used as generators in design exploration workflows.

\subsection{Challenges}\label{sec:challenges}

While Architext models have produced very promising results for Architext models, we are aware that there are a number of limitations and challenges in the current generative pipeline and experimental protocol, which we list below.

\paragraph{Specificity of the task} \space While Architext achieves impressive performance in generating residential layouts out of natural language prompts, it is also limited to generating specifically these type of layouts, trained on a narrow range of semantic annotations. This specialization partly explains the impressive performance, especially in some categories of prompts. Allowing for a larger diversity of designs and semantic annotations would go a long way to indicate the resilience and robustness of this approach. Additionally, the designs generated do not include many details that might be useful for downstream tasks (e.g. entrances, doors, balconies, etc.), limiting Architext currently to the first step of conceptual design (i.e. ideation on top of morphology and structure).

\paragraph{Limited range of constraints} \space Architext is trained to consider a limited number of constraints, specifically location, room numbers, and adjacency relationships. While these constraints alone allow for a large variety of designs to be specified, they also constrain the type of conditioning we can put on the generated designs and the range of outputs Architext generates.

\paragraph{Auto-regressive models} \space
While auto-regressive language models have achieved impressive performance across a variety of domains and tasks, there is a number of limitations that their architecture imposes. First, left-to-right prediction does not allow the model to utilize information in partly generated prompts, making tasks such as space in-filling impossible---unlike bidirectional models \cite{Liu2019RoBERTaAR} which can handle such tasks with ease. Second, auto-regressive models can not easily tackle multi-task training \cite{Raffel2020ExploringTL} (i.e. training the model with more than one ways to generate an output). Collectively, such limitations make it harder to train Architext models on different design tasks at once, with a variety of goals (and even different types of data), which define a typical real world design setting.

\paragraph{Scaling across tasks} \space

There is a large number of important and interesting design tasks available within the AEC industry. Finetuning models across all these different tasks, and making sure models stay up to date with new requirements and constraints that might be imposed, is an extremely costly process. Exploring efficient methods of finetuning is a prerequisite to implementing design workflows driven by such models. To handle multi-task learning, it would be necessary to expand Architext with new learning methods such as in-context training \cite{Kojima2022LargeLM} (without retraining the model) or the use of adapters \cite{Hu2021LoRALA} to finetune the model to a new task.

\subsection{Future Work}\label{sec:future_work}

Architext offers an exciting new paradigm for design. Semantic generation, and more broadly large-scale language modelling, opens many interesting avenues for future experimentation. A few such ideas are described below. This is not meant to be an exhaustive list but rather initial stepping stones in avenues of research that we think are important.

\paragraph{Finetuning with external rewards} \space
Given the structured nature of the generated outputs and the ease with which we can evaluate their performance across many different metrics, finetuning the model with such signals as rewards appears to be a promising study for improving model performance. Performance metrics could be tied with the model's generation capacity (e.g. validity, correctness, or diversity of outputs) or with performance metrics that relate to real-world aspects of designs (e.g. daylight, energy demand, comfort, etc.). Additionally, this approach would allow us to combine offline training of these models, on top of real or synthetic datasets, with online training on top of human designer decisions. A Human-In-The-Loop implementation that uses such designer preferences to finetune a trained model would be a very promising future extension of Architext, in the vein of \cite{Christiano2017DeepRL}.

\paragraph{Human-Model Interaction} \space
Another promising avenue of future research is to utilize the means within which we can interpret and evaluate a language model when deployed alongside a human designer. One such method would be to teach the model to express its own uncertainty with respect to a certain part of its generation, and do it in a way that is easily visible to the human designer \cite{Guo2021LearningTC}. This would provide an easy way for the human designer to intervene in the design generation process and interact with the model. Such interactions could also become important data points for external rewards used to further finetune these models, as described earlier. A much needed research extension would be to deploy a system like Architext in production, within a real-world design workflow where practitioners can use it in their day-to-day tasks. Such human user studies are crucial for evaluating not only the capacity of the model in providing positive input to the design process but also the ability and potential difficulty of designing within such a novel design workflow. The user experience lessons learned from such integration would be invaluable for the future design of such systems.

\paragraph{Design as sequence of decisions} \space
Another potential avenue of future research would be to think of design as a sequential process of decisions. In this way, generating a design artefact would be akin to planning, allowing the model to generate a step-by-step process of design \cite{Nye2021ShowYW,Huang2022LanguageMA}. This enables new ways to utilize the model as an agent within environments of experience (e.g. design exploration environments) where the model can now take specific decisions towards designing outputs. This approach can be further extended by utilizing an upside down Reinforcement Learning process \cite{Schmidhuber2019ReinforcementLU} where the model can once again be trained on offline datasets of semantic annotations and designs, only this time each design can be accompanied by a specific `reward-to-go', a quantification of a certain design performance metric. Trained in this way, these Decision Transformer models \cite{janner2021offline,chen2021decision} would be able to generate designs based on specific preferences with respect to performance. This upside down approach would revolutionize design which traditionally revolves around generating a design and then evaluating its performance, versus creating a design \textit{conditioned} on performance.

\paragraph{Expanding the scope} \space
A final avenue of research is of course to expand the scope of applications that are tackled. The way Architext operationalizes semantic generation is not specific to design but can easily be generalized across different tasks, as long as it is possible to provide semantic annotations for both inputs and outputs. New tasks could include almost any generative task the AEC deals with, such as allocating structural columns within a floorplate, designing a new district or masterplan, detailing landscape design on top of a site, and more. Of course, expanding the scope is not limited on finding different tasks but also similar tasks in adjacent domains. One promising avenue of research would be applying Architext for procedural content generation within games \cite{togelius2010sbpcg}, driven by experience \cite{Yannakakis2011} or within powerful exploration frameworks such as Quality Diversity (QD) \cite{Gravina2019ProceduralCG}. The intersection with QD is a research area of particular importance for Architext given its ability to easily generate a large diversity of layouts, acting as a powerful generator for QD-driven, design exploration workflows. Substituting combinatorial search processes (that suffer from the curse of dimensionality) with pre-trained language models that can be deployed and sampled easily, would enable a wide range of design exploration applications. 

\section{Conclusion}\label{sec:conclusion}

Motivated by the potential and increasing availability of highly capable pretrained language models, we studied the potential of architectural design generation via the use of language, and specifically a geometric representation of design conveyed through language. We showed that pretrained language models can be fine-tuned into architectural design generators that can, in turn, produce a large number of valid and diverse residential layouts. This study introduces the tool that is able to perform this task, Architext, and with it a new research area at the intersection of language and design. Architext offers the possibility of utilizing efficient intermediate representations such as language, geometry, and code, to train and deploy powerful generative models for practically any design task.

Our results suggest that scaling the size of the language model provides a boost in semantic accuracy, without affecting design diversity. Such a finding indicates that better performing models are within reach. A full scaling laws study \cite{kaplan2020scaling} for architectural design capability is an open area of research for this work, evaluating both model and data size scaling and their relationship to semantic accuracy and design diversity. It is possible that newer, larger, optimally trained language models can enable powerful downstream capabilities and drive useful applications in practice. For instance, Quality Diversity search \cite{cully2018qd} could help Architext create larger and more diverse datasets that can then be used to train better performing models, opening the way for an open-ended, iterative, improvement process.


\bibliographystyle{unsrt}
\bibliography{bibliography}

\appendix
\newpage
\section{Prompts used for generating designs from fine-tuned models}\label{sec:appendix}

\begin{center}
\begin{longtable}{ |l|l|l| } 
 \hline
 \textbf{ID} & \textbf{Prompt} & \textbf{Geometric property} \\
 \hline
AN.1 & the bedroom is not adjacent to the living room & negative adjacency \\
AN.2 & a bedroom is not adjacent to the living room & negative adjacency \\
AN.3 & the bedroom is not adjacent to the kitchen & negative adjacency \\
AN.4 & a bedroom is not adjacent to the kitchen & negative adjacency \\
AN.5 & the bedroom is not adjacent to the kitchen & negative adjacency \\  
AN.6 & the kitchen is not adjacent to the bathroom & negative adjacency \\
AN.7 & a bathroom is not adjacent to the living room & negative adjacency \\
AN.8 & the bathroom is not adjacent to the living room & negative adjacency \\
 \hline
  \hline
AP.1 & the bedroom is adjacent to the living room & positive adjacency \\
AP.2 & a bedroom is adjacent to the living room & positive adjacency \\
AP.3 & the bedroom is adjacent to the kitchen & positive adjacency \\
AP.4 & a bedroom is adjacent to the kitchen & positive adjacency \\
AP.5 & the bedroom is adjacent to the kitchen & positive adjacency \\  
AP.6 & the kitchen is adjacent to the bathroom & positive adjacency \\
AP.7 & a bathroom is adjacent to the living room & positive adjacency \\
AP.8 & the bathroom is adjacent to the living room & positive adjacency \\
 \hline
  \hline
LNU.1 & a bedroom is in the north side of the house & non-unique location \\
LNU.2 & a bedroom is in the north east side of the house & non-unique location \\
LNU.3 & a bedroom is in the east side of the house & non-unique location \\
LNU.4 & a bedroom is in the south east side of the house & non-unique location \\
LNU.5 & a bedroom is in the south side of the house & non-unique location \\
LNU.6 & a bedroom is in the south west side of the house & non-unique location \\
LNU.7 & a bedroom is in the west side of the house & non-unique location \\
LNU.8 & a bedroom is in the north west side of the house & non-unique location \\
\hline
\hline
LU.1 & the bedroom is in the north side of the house & unique location \\
LU.2 & the bedroom is in the north east side of the house & unique location \\
LU.3 & the bedroom is in the east side of the house & unique location \\
LU.4 & the bedroom is in the south east side of the house & unique location \\
LU.5 & the bedroom is in the south side of the house & unique location \\
LU.6 & the bedroom is in the south west side of the house & unique location \\
LU.7 & the bedroom is in the west east side of the house & unique location \\
LU.8 & the bedroom is in the north west side of the house & unique location \\
LU.9 & the kitchen is in the north side of the house & non-unique location \\
LU.10 & the kitchen is in the north east side of the house & non-unique location \\
LU.11 & the kitchen is in the east side of the house & non-unique location \\
LU.12 & the kitchen is in the south east side of the house & non-unique location \\
LU.13 & the kitchen is in the south side of the house & non-unique location \\
LU.14 & the kitchen is in the south west side of the house & non-unique location \\
LU.15 & the kitchen is in the west east side of the house & non-unique location \\
LU.16 & the kitchen is in the north west side of the house & non-unique location \\
 \hline
  \hline
RG.1 & a house with five rooms & \# of rooms (generic) \\
RG.2 & a house with six rooms & \# of rooms (generic) \\
RG.3 & a house with seven rooms & \# of rooms (generic) \\
RG.4 & a house with eight rooms & \# of rooms (generic) \\
RG.5 & a house with nine rooms & \# of rooms (generic) \\
RG.6 & a house with ten rooms & \# of rooms (generic) \\
 \hline
 \hline
RS.1 & a house with one bedroom and one bathroom & \# of rooms (specific) \\
RS.2* & a house with one bedroom and two bathrooms & \# of rooms (specific) \\
RS.3 & a house with two bedrooms and one bathrooms & \# of rooms (specific) \\
RS.4 & a house with two bedrooms and two bathrooms & \# of rooms (specific) \\
RS.5 & a house with two bedrooms and three bathrooms & \# of rooms (specific) \\
RS.6 & a house with three bedrooms and one bathroom & \# of rooms (specific) \\
RS.7 & a house with three bedrooms and two bathrooms & \# of rooms (specific) \\
RS.8* & a house with three bedroom and three bathrooms & \# of rooms (specific) \\
RS.9* & a house with four bedrooms and one bathroom & \# of rooms (specific) \\
RS.10 & a house with four bedrooms and two bathrooms & \# of rooms (specific) \\
RS.11 & a house with four bedrooms and three bathrooms & \# of rooms (specific) \\
RS.12* & a house with four bedrooms and four bathrooms & \# of rooms (specific) \\
 \hline

\end{longtable}
\end{center}

\end{document}